# Enhancing Prosthetic Safety and Environmental Adaptability: A Visual-Inertial Prosthesis Motion Estimation Approach on Uneven Terrains

Chuheng Chen, Xinxing Chen, Shucong Yin, Yuxuan Wang, Binxin Huang, Yuquan Leng and Chenglong Fu

*Abstract*—Environment awareness is crucial for enhancing walking safety and stability of amputee wearing powered prosthesis when crossing uneven terrains such as stairs and obstacles. However, existing environmental perception systems for prosthesis only provide terrain types and corresponding parameters, which fails to prevent potential collisions when crossing uneven terrains and may lead to falls and other severe consequences. In this paper, a visual-inertial motion estimation approach is proposed for prosthesis to perceive its movement and the changes of spatial relationship between the prosthesis and uneven terrain when traversing them. To achieve this, we estimate the knee motion by utilizing a depth camera to perceive the environment and align feature points extracted from stairs and obstacles. Subsequently, an error-state Kalman filter is incorporated to fuse the inertial data into visual estimations to reduce the feature extraction error and obtain a more robust estimation. The motion of prosthetic joint and toe are derived using the prosthesis model parameters. Experiment conducted on our collected dataset and stair walking trials with a powered prosthesis shows that the proposed method can accurately tracking the motion of the human leg and prosthesis with an average root-mean-square error of toe trajectory less than 5 cm. The proposed method is expected to enable the environmental adaptive control for prosthesis, thereby enhancing amputee's safety and mobility in uneven terrains.

## I. INTRODUCTION

The emergence of powered prosthesis has made a great contribution to enabling millions of amputees to regain their walking abilities [1]-[3]. However, significant challenges remain achieving safe and stable locomotion with prosthesis on uneven terrains such as stairs and obstacles [4]. Inspired by the human use of vision to guide their own movement in nonrhythmic locomotion in complex environments [5], the integration of environmental information captured by wearable vision sensors has shown potential in enhancing prosthetic locomotion across various terrains [6], [7].

Researchers have explored the use of different vision sensors in conjunction with classification algorithms to accomplish terrain classification (e.g., stair, ramp and level ground), as well as the estimation of corresponding parameters, such as stair height and width [8]-[10]. This information provides valuable environmental context that facilitates the seamless transition between different locomotion modes of the prosthesis [11]. However, current environmental perception system for prosthesis only provides terrain types and parameters, which falls short of fully leveraging the environmental information gathered from vision sensors. The inconsistence between the classification results and the amputee's real locomotion mode, and limited awareness of the prosthesis's movement and its spatial relationship with uneven elements of terrains may lead to collisions and falls [12]. In addition, the terrain parameters are calculated only based on the information of current frame, thus the estimation result may be incorrect when complete terrain information (e.g., a complete step of stairs) is not available. Inconsistencies existing in the parameters between different steps may also reduce the foot clearance and pose risks for amputee's stair and obstacle negotiation [13]. Therefore, navigating complex terrains, like stairs, obstacles, and transitions between different terrains, still poses risks for amputees.

To enhance safe and predictive walking on uneven and unstructured terrains, our goal is to comprehensively utilize environmental information to estimate the prosthesis's movement within the environment. While visual odometry and SLAM are commonly used for motion estimation in mobile robots [14], [15], there is a dearth of research focusing on motion estimation for prosthesis within the environment. Besides, researchers have developed obstacle location and tracking system for exoskeleton and adaptively changed the step length for obstacle crossing [16]. However, different from the biped robot [17] and exoskeleton [16], planning a global trajectory for prosthesis to avoid the collision with uneven terrains may not be suitable. This is because the prosthesis motion depends on the motion of the amputee, making it impractical to follow a pre-planned trajectory. In addition, estimating the prosthesis's motion state requires mounting a vision sensor on it, leading to challenges like limited camera field of view, reduced texture information, and rapid view angle changes. Therefore, the vision sensors may only detect part of the obstacle and stair risers due to the constrained field of view. Moreover, in such

This work was supported by the National Natural Science Foundation of China [Grant U1913205, 62103180]; Guangdong Innovative and Entrepreneurial Research Team Program [Grant 2016ZT06G587]; the Stable Support Plan Program of Shenzhen Natural Science Fund [Grant 20200925174640002]; the Science, Technology and Innovation Commission of Shenzhen Municipality [Grant SGLH20180619172011638, ZDSYS20200811143601004, KYTDPT20181011104007 and JCYJ20230807093407001]; and Centers for Mechanical Engineering Research and Education at MIT and SUSTech. (Corresponding author: Chenglong Fu.)

The authors are all with Shenzhen Key Laboratory of Biomimetic Robotics and Intelligent Systems, Shenzhen, 518055, China, and also with Guangdong Provincial Key Laboratory of Human-Augmentation and Rehabilitation Robotics in Universities, Southern University of Science and Technology, Shenzhen, 518055, China (email: fucl@sustech.edu.cn and chenxx@sustech.edu.cn).

situations, feature-based methods like ORB-SLAM [18] may struggle to extract sufficient feature points from grayscale images, resulting in tracking loss. Typical RGB-D cameras, such as Intel RealSense and Microsoft Kinect [19], are also too large for prosthesis integration [10]. Besides, real-time integration with prosthesis control necessitates swift motion estimation calculations, making 3D environmental point cloud-based methods impractical due to their computational demands [20]. Therefore, designing a lightweight motion estimation method tailored to the prosthesis's motion characteristics is imperative.

To overcome these limitations and enhance the environmental awareness and self-awareness of the prosthesis, a visual-inertial motion estimation approach is proposed in this paper. The proposed approach utilizes the environmental point cloud captured by a depth camera for motion estimation of the prosthesis in the sagittal plane, aiming to enhance adaptative walking on uneven terrains. The sagittal plane's environmental structure significantly impacts users' gait and prosthetic control, making it the central focus of our study [21]. To that end, feature points are extracted from 2D environmental point cloud of stairs and obstacles and then used for frame-to-frame alignment through an iterative closest-point (ICP) approach to obtain an initial camera motion estimation. Moreover, to improve the robustness and accuracy in camera motion estimation, a sagittal plane motion-oriented error-state Kalman filter (SP-ESKF) is incorporated to fuse the visual and inertial estimation of the camera motion in the sagittal plane. Finally, the camera motion is combined with the prosthesis model to derive the movement of each joint and toe of the prosthesis.

The primary contributions of the present paper include:

1) Developing a visual-inertial motion estimation method for prosthesis in uneven terrains to estimate its movement and position within the environment, enhancing the environmental adaptability and walking safety for prosthetic user.

2) Proposing a sagittal plane motion-oriented error-state Kalman filter with a reduced parameter set, facilitating real-time environmental adaptive prosthesis control.

3) Evaluating the proposed motion estimation approach using our collected dataset on different terrains and in a stair ascent walking experiment wearing a powered prosthesis.

The rest of this paper is described as follows. Section II introduces the proposed visual-inertial motion estimation approach for prosthesis. The experiment setup and results are presented in Section III and discussed in Section IV. Section V is the conclusion of this paper.

## II. METHODS

The proposed visual-inertial motion estimation approach is introduced in this section. An overview of this approach is presented in Fig. 1. The proposed method consists of three main components: visual estimation, inertial navigation and information fusion for motion estimation. First, the preprocessed environmental point cloud captured by a depth camera fixed on prosthetic knee is utilized to initially estimate the knee motion through frame-to-frame alignment with ICP algorithm. After accomplishing the estimation of knee motion with inertial information, the estimation results are fused with SP-ESKF and the position of the prosthesis is derived by incorporating the prothesis model parameters.

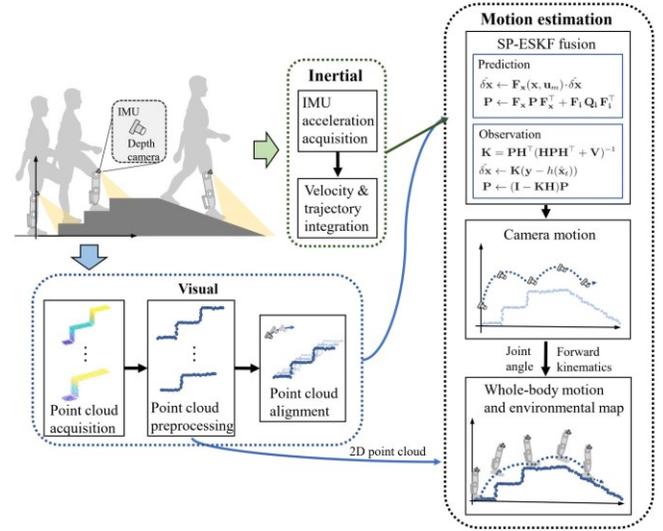

Fig. 1. The overview of the proposed visual-inertial prosthesis motion estimation approach. The visual estimation of camera and prosthetic knee motion using ICP algorithm and the inertial estimation are fused with SP-ESKF to obtain the final motion estimation. The estimation results are utilized to derive the whole-body motion of the prosthesis with joint angle and forward kinematics.

*A. Visual Estimation Based on ICP*

*1) Environmental Data Preprocessing*

The 3D environmental point cloud in front of the user is acquired with a depth camera (CamBoard pico flexx2, pmdtechnologies, Germany) fixed on the knee of the prosthesis. The relative motion of the camera and the prosthesis in the environment is consistent due to their solid connection. Before alignment, preprocessing of the point cloud is required, which consists of two parts: sagittal plane projection and coordinate system conversion.

The point cloud directly acquired by the depth camera contains 3D coordinate information of about 38k points per frame. Alignment for these points imposes a huge computational burden, which may present difficulties in integrating with real-time control of the prosthesis. As stated above, this study mainly focuses on the environmental information and the prosthesis motion in the sagittal plane. To reduce the computational effort, the 3D point cloud is projected into the sagittal plane for dimension reduction as:

$$\begin{cases} \boldsymbol{P}_{3D} = \{(x_i, y_i, z_i) \mid i = 1, ..., n\} \\ \boldsymbol{J} = \{i \mid -0.05\text{m} < y_i < 0.05m\}, \\ \boldsymbol{P}_{2D} = \{(x_j, z_j) \mid j \in \boldsymbol{J}\} \end{cases} \quad (1)$$

where $\boldsymbol{P}_{3D}$ and $\boldsymbol{P}_{2D}$ are original 3D environmental point cloud and point cloud in the sagittal plane respectively. $x_i$, $y_i$, and $z_i$ are the coordinate values of the point. $\boldsymbol{J}$ is the index set of points extracted from the 3D point cloud directly in front of the prosthesis. With sagittal plane projection, the original 3D point cloud is converted into 2D point cloud in the sagittal plane, and meanwhile, the camera motion can be considered as motion in the sagittal plane, which reduces the computation time.

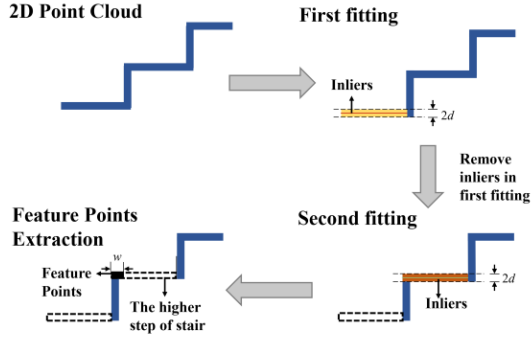

Fig. 2. The feature points extraction proscess. The yellow and brown lines are inliers fitted in the first and second fitting. The black short line incorporates the extracted feature points.

In addition, considering that the main concern of the study is to utilize the camera position to derive the prosthesis motion, an inertial measurement unit (IMU) (MTi 1-series, Xsens, Netherlands) is used to fixed with the camera to acquire the orientation information of the camera and convert the coordinate system of the point cloud from the camera system to the ground system. Therefore, the camera motion can be considered as translational motion, and the coordinate system is converted as:

$$P_{Gnd} = R_{Cam}^{Gnd} P_{Cam}, \quad (2)$$

where $P_{Gnd}$ and $P_{Cam}$ denote the point cloud of the ground and camera coordinate systems respectively. $R_{Cam}^{Gnd}$ describes the rotation matrix from the camera coordinate system to the ground coordinate system calculated with the Euler angle of IMU. Besides, to reduce the impact of noise due to camera performance or ground material on the accuracy of alignment, the point cloud is smoothed using the k-Nearest Neighbor algorithm at last.

*2) Feature Point Extraction*

After pre-processing, the original 3D environmental point cloud is converted to 2D point cloud in the ground coordinate system, and the camera motion is simplified to translational motion in the sagittal plane. However, alignment of the whole point cloud containing redundant points may lead to long alignment time and increased errors. To overcome this, researchers have extracted feature points in the point cloud before alignment [22]. Since the aim of this study is to help the predictive walking in uneven terrain such as stairs and obstacles, considering the characteristics of these terrains, the corner points of stairs and obstacles are manually extracted using the random sample consensus (RANSAC) algorithm and used for subsequent frame-to-frame alignment.

Before feature point extraction, the terrain in front of the user is classified using the CNN-based environmental recognition method proposed in our previous work [9]. If the terrain is recognized as stair or ground with obstacle, the RANSAC algorithm is used to extract the two median lines of the stair or the obstacle.

As shown in Fig. 2, a horizontal line with the most inliers is fitted first. After removing the inliers of the first line, another horizontal line with the most inliers is fitted. The median z coordinates of the two lines are compared and the feature points are extracted from the end of the line with larger z-coordinate (the top of the obstacle or the upper level of the staircase), where $d$ and $w$ are manually set thresholds of 0.02 m and 0.05 m, respectively.

*3) Frame-to-frame Alignment with ICP*

The movement of the camera can be estimated by frame-to-frame alignment of the feature points extracted from the point cloud. After extraction, if difference between the mean coordinate values of the feature points extracted at two consecutive moments $t_i$ and $t_{i-1}$ is within a preset threshold, the feature points of two frames are considered to be the same corner of the terrain. Then, frame-to-frame alignment is realized using ICP algorithm [23] and the displacement $T$ between the point cloud in the user's sagittal plane is estimated. The ICP workflow is described as follows:

1) Given a source point cloud $P_s$, a target point cloud $P_t$ and an initial prediction $T_0$ of the transformation between $P_s$ and $P_t$, for each point in $P_t$ ($P_i \in P_t$), corresponding point that has the closest distance to it in the initially transformed source point cloud is searched:

$$P_i' = P_j, \; j = \arg\min_j \left\| P_i - (T + P_j) \right\|, P_j \in P_s. \quad (3)$$

2) The displacement between point clouds are estimated using corresponding point pairs to minimize the error and then injected into the displacement estimated in the previous step:

$$\begin{cases} T^{n+1} = \triangle T + T^n \\ \triangle T = \arg\min_{\triangle T} \left\| P - (\triangle T + P') \right\| \end{cases}. \quad (4)$$

3) The above steps are applied iteratively until converging ($\triangle T \leq T_{th}$, $T_{th}$ is taken as $10^{-6}$) or the maximum number of iterations (taken as 20) is reached to obtain the final displacement.

*B. IMU Measurement Model*

The initial motion estimation of camera and prosthetic knee is achieved by frame-to-frame alignment. However, tracking errors caused by incorrect extraction of feature points or inability to extract feature points still exist. To solve this problem, the camera's inertial information measured by IMU can be used to estimate the camera motion state.

IMU is capable of measuring three-axis acceleration and angular velocity, which are affected by Gaussian white noise and zero bias. Since the camera motion has been considered as translational motion, the motion equations regarding angular velocity is not considered in this study. Therefore, the acceleration of IMU can be described by the following model:

$$\begin{cases} a_m(t) = R^{-1}\left(a_w(t) - g\right) + a_b(t) + a_n(t) \\ a_n \sim N(0, \sigma_a^2) \\ \dot{a}_b = a_{bn} \sim N(0, \sigma_{a_b}^2) \end{cases}, \quad (5)$$

where $a_m$ and $a_w$ represent measured acceleration and acceleration in the world coordinate system; $a_n$ and $a_b$ denote accelerometer noises following the Gaussian distributions and accelerometer bias whose derivatives follow the Gaussian distributions and $R$ is the rotation matrix from the IMU coordinate system to the ground coordinate system in SO(2) calculated with the IMU Euler angle. Besides, $a_n$ and $a_b$ are given by the IMU vendor, Xsens Technologies.

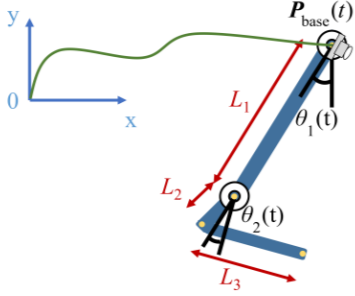

Fig. 3. The link model of the prosthesis. $P_{base}$ denotes position of the knee estimated by the motion of camera. $\theta_1$ and $\theta_2$ denote the angle calculated using the joint angle measure by two encoders fixed on the knee and ankle of the prosthesis.

By integrating the acceleration, velocity and displacement of the IMU can be obtained. However, due to the noise and bias, direct integration can lead to error accumulation. To avoid the error increasing with time, it is also necessary to fuse the visual information to rectify the integration results.

*C. Visual-inertial Fusion Based on SP-ESKF*

ESKF combines the efficiency of Kalman filter with the flexibility of error state estimation and has been used in navigation and positioning systems. Since our main concern is the motion estimation of the knee in the sagittal plane and inspired by [24], a sagittal plane motion-oriented error state Kalman filter with a reduced set of parameters is proposed to combine the visual with the inertial estimation results to obtain a more robust and accurate knee motion estimation. The total system state of the knee motion is described by the true state $x_t = (p_t, v_t, a_{bt}, g_t)$, nominal state $x = (p, v, a_b, g)$ and error state $\delta x = (\delta p, \delta v, \delta a_b, \delta g)$.

The workflow can be divided into two steps: 1) updating the nominal state from the IMU measurement results, and updating the error state and covariance matrix according to the error state processing model; 2) calculating the Kalman gain K and rectifying the error state using the visual estimation results, and finally injecting the error state into the nominal state to obtain the true state of the system.

*1) Updating Process of Nominal State and Error State*

When IMU data is fed in, the nominal state is integrated and provide a prediction of knee movement, and meanwhile, the error state and covariance matrix will be updated. Since the system error is concentrated into the error state, the updating process of the nominal state can be described as:

$$\begin{cases} p \leftarrow p + v\Delta t + \frac{1}{2}(R(a_m - a_b) + g)\Delta t^2 \\ v \leftarrow v + (R(a_m - a_b) + g)\Delta t \\ a_b \leftarrow a_b \\ g \leftarrow g \end{cases}. \quad (6)$$

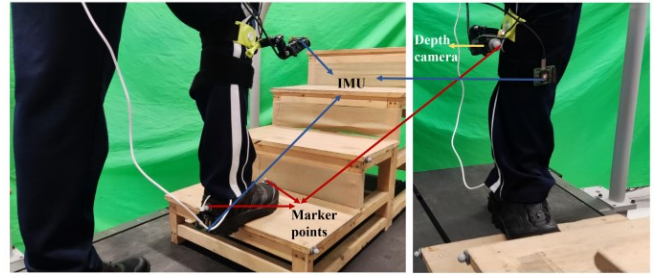

(a) The environment setup for data collection

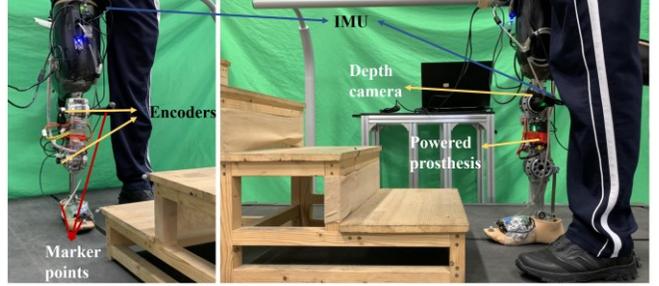

(b) The environment setup for prosthesis walking trails

Fig. 4. The experimental environment of our collected dataset and prosthesis walking. In our collected dataset of walking in different terrains, the depth camera and IMU are mounted on the shank of the subject. In prosthesis walking experiment, the depth camera and IMU are mounted on the knee of the prosthesis.

The updating process in matrix form of the error state is described as:

$$\begin{cases} \delta x \leftarrow F_x \delta x \\ P \leftarrow F_x \cdot P \cdot F_x^T + F_i \cdot Q \cdot F_i^T \\ F_x = \begin{bmatrix} I & F_1 & 0 & 0 \\ 0 & I & F_2 & F_1 \\ 0 & 0 & I & 0 \\ 0 & 0 & 0 & I \end{bmatrix} \in \mathbb{R}_{8 \times 8} \\ F_i = \begin{bmatrix} 0 & 0 \\ I & 0 \\ 0 & I \\ 0 & 0 \end{bmatrix} \in \mathbb{R}_{8 \times 4}, \delta x = \begin{bmatrix} \delta p \\ \delta v \\ \delta a_b \\ \delta g \end{bmatrix} \in \mathbb{R}_{8 \times 1} \\ Q = \begin{bmatrix} \sigma_a^2 \Delta t^2 I & 0 \\ 0 & \sigma_{a_b}^2 \Delta t I \end{bmatrix} \in \mathbb{R}_{4 \times 4} \\ F_1 = I\Delta t, F_2 = -R\Delta t \end{cases}, \quad (7)$$

where $P$ and $Q$ denote the covariance matrix of the error state and the perturbation input; $F_x$ is system transition matrix and $F_i$ is perturbation transition matrix.

*2) Error State Correction and Data Fusion*

When the frame-to-frame alignment is completed, the output of alignment will serve as a measurement to correct the error state and achieve the fusion of visual and inertial estimation. Calculation of Kalman gain $K$ and the correction of error states and the covariance matrix can be described as:

$$\begin{cases} \boldsymbol{K} = \boldsymbol{PH}^{\mathrm{T}}(\boldsymbol{HPH}^{\mathrm{T}} + \boldsymbol{V})^{-1} \\ \delta \boldsymbol{x} \leftarrow \boldsymbol{K}(\boldsymbol{z} - h(\boldsymbol{x}_t)) \\ \boldsymbol{P} \leftarrow (\boldsymbol{I}_{8\times8} - \boldsymbol{KH})\boldsymbol{P} \\ \boldsymbol{H} = \dfrac{\partial h}{\partial \delta \boldsymbol{x}} = [\boldsymbol{I}_{2\times2}\ \ 0\ \ 0\ \ 0] \in \mathbb{R}_{2\times8} \\ \boldsymbol{z} = h(\boldsymbol{x}_t) + \boldsymbol{v}, \boldsymbol{v} \sim N(0, \boldsymbol{V}) \end{cases}, \quad (8)$$

where $\boldsymbol{P}$ represents Jacobian matrix of the observation $h(x)$ with respect to the error state and $\boldsymbol{V}$ represents the covariance matrix of the observation noise.

Then, the corrected error state is injected into the nominal state to obtain the truth state of the system, and finally, the error state will be reset to zero and the covariance matrix is updated with the newest nominal state.

*D. Prosthesis Motion Estimation and Control Scheme*

After achieving the estimation of the knee motion, the movement of the joint and toe of the prosthesis are derived with the prosthesis link model and forward kinematics.

As shown in Fig. 3, the base of the prosthesis is a removable knee joint and the end (prosthesis foot) is a bent linkage with a constant angle. By fixing another IMU to user's thigh, the angle between the thigh and the coronal plane can be measured and the clockwise rotation angle of the knee and ankle joints can be obtained with two encoders fixed on the two joints. Then, the positions of the ankle, heel and toe in the ground coordinate system can be calculated with forward kinematics. In addition, after obtaining the camera motion, the key frames are extracted based in the camera displacement, and the 2D point clouds of the key frames are added to the local map to construct the 2D point cloud environment map. In this study, the frames with displacements between consecutive frames within the range of 0.01 m to 0.05 m are taken as key frames.

After realizing the motion estimation, to visually demonstrate the effect of the proposed method for prosthetic walking on uneven terrain, we propose an upstair walking control scheme for the prosthesis and integrate the estimated prosthesis motion into the scheme. First, an IMU installed on the socket of the prosthesis and a six-dimensional force sensor at the ankle are utilized to measure the angle between the thigh and the coronal plane of the body, as well as the ground reaction force. These measurements determine the status of the human leg (flexion/extension) and prosthesis (stance/swing phase). In the swing phase, we iteratively compute the optimized joint angles within the constraint range, employing multi-objective optimization that considers obstacle avoidance, biomimicry, smooth landing, and motion fluidity. This iterative process involves seeking the minimum value of the cumulative optimization objective function of the prosthesis motion state across sequential time steps to predictively derive the desired joint angles. Subsequently, the prosthesis is actuated using a position-based lower-level PD control to follow the desired joint angles. Since the prosthesis control scheme is not the mainly focus of this paper, the detailed strategy will be described in our future work.

## III. EXPERIMENT AND RESULTS

*A. Experimental Protocol*

To evaluate the proposed visual-inertial motion estimation method, an able-bodied subject was invited to perform 5 times of walking under three terrains (upstair, downstair and obstacle) to collect environmental point cloud and IMU data to estimate the motion of camera and the subject's foot. The experiment environment is shown in Fig. 4(a). The height and width for stair and obstacle are 14.7 cm and 28 cm (for stair) and 14 cm and 13 cm (for obstacle), respectively. The depth camera and IMU were mounted together on subject's right shank, and another two IMUs were mounted on the subject's right shank and heel to obtain the ankle angle. The capture frame rate of the point cloud and the frequency of the IMUs were 30 frames per second and 100 Hz, respectively. Visual and IMU data are collected in two threads and time-synchronized by capturing and fusing the latest data from both threads. In addition, in the experiments, a motion capture system (Raptor-12HS, Motion Analysis Corporation, USA) was used to capture the motion of the subject's lower limbs and the camera at a frequency of 120 Hz. The measured values of the motion capture system were used as the ground truth of the camera and lower limb motion, and the proposed method was evaluated by calculating the absolute trajectory error (ATE) between the motion trajectory estimated by the propose method and measured by the motion capture system. Although the prosthesis motion between each frame is used in our prosthesis control scheme instead of ATE, it can still demonstrate that the proposed method can realize accurate motion estimation.

In addition, to evaluate the proposed method on powered prosthesis, as shown in Fig. 4(b), the subject was also invited to perform upstair walking experiment wearing a powered prosthesis. The prosthesis follows the control scheme mentioned in Section II.D. The same depth camera and an IMU are fixed together on the knee of the prosthesis to estimate its motion and the motion of the prosthesis foot is calculated with the joint angle acquired by two encoders mounted on knee and ankle joint of the prosthesis. The motion capture system was also used to obtain the truth of the prosthesis motion. Besides, in all experiments, the subject began walking in a position close to stair or obstacle and started performing the corresponding gait directly.

The experimental data processing was based on Python 3.7 and Matlab 2020b and performed on a computer with an AMD Ryzen 7 4800H CPU (2.9 GHz), 16 GB of RAM, and an NVIDIA Geforce RTX 2060 GPU.

*B. Estimation Result on the Collected Dataset*

The proposed method is firstly evaluated using the dataset collected from healthy subjects walking in different terrains. In each frame, the average time cost of point cloud preprocessing and visual estimation is 6 ms and 19 ms respectively. In addition, the time cost of IMU integration, data fusion, prosthesis motion calculation per frame are all below 1 ms. Therefore, the total forward time is shorter than the acquisition time of the depth camera ($>$ 30 ms), showing the promising real-time performance of the proposed method.

The estimated toe position of the subject in sagittal plane and the ground truth during walking in different terrains are shown in Fig. 5(a)-Fig. 5(c) and the motion of the subject's shank and foot and the constructed point cloud map during walking are shown in Fig. 5(a)- Fig. 5(c). The trajectories show that the proposed method tracks the toe motion well in all three uneven terrains. The average ATE in upstair and downstair walking are 2.96 cm ± 0.92 cm (upstair, x), 2.16 cm ± 0.46 cm (upstair, z), 2.79 cm ± 0.79 cm (downstair, x) and 1.63 cm ± 0.23 cm (downstair, z), respectively. In obstacles, since the camera is not able to see the obstacle during the whole crossing, only the periods when the obstacle is visible are taken for ATE calculation, and the average ATE are 2.20 cm ± 0.14 cm (x) and 4.68 cm ± 0.47 cm (z), respectively, which are still less than 5 cm.

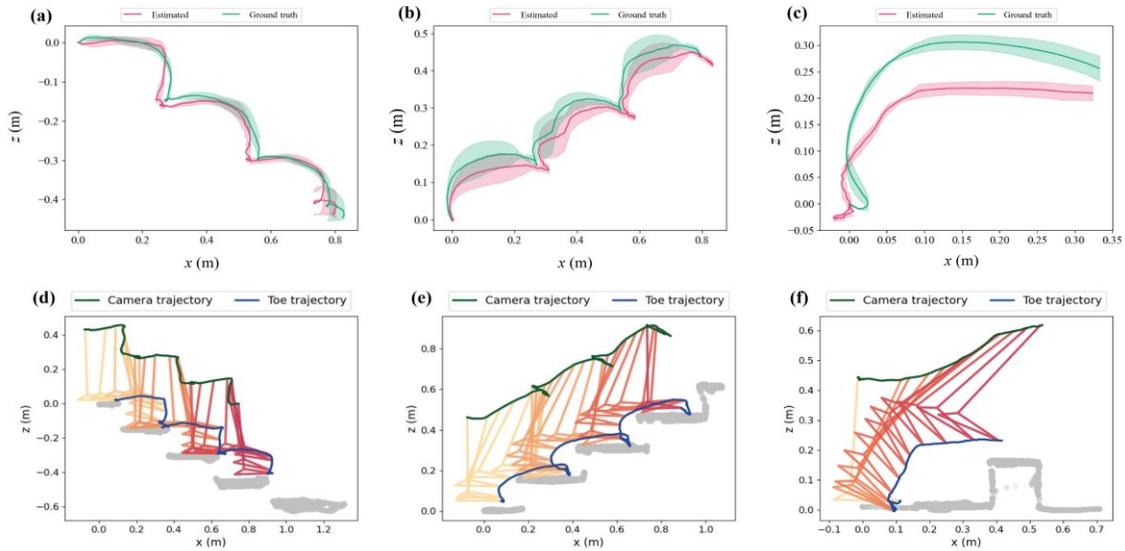

Fig. 5. The experiment result on the collected dataset. (a), (b) and (c) describe the estimated toe position of the subject in sagittal plane and the ground truth in upstair, downstair and obstacle. The solid line and the shading in the graph represent the mean ± standard deviation of the toe positions, respectively. (d), (e) and (f) describe the estimated motion of the subject's leg in sagittal plane and the envirnional map of 2D point cloud.

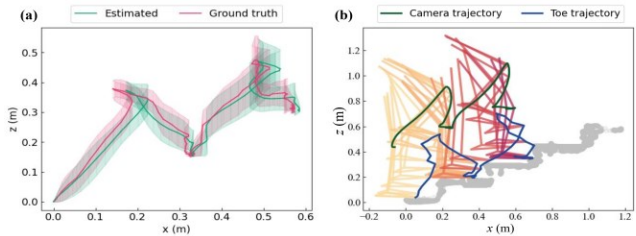

Fig. 6. The experiment result on data collected during prosthesis upstair walking. (a) describes the estimated toe position of the subject in sagittal plane and the ground truth. The solid line and the shading in the graph represent the mean ± standard deviation of the toe positions, respectively. (b) describes the estimated motion of the prosthesis in sagittal plane and the envirnional map of 2D point cloud.

*C. Estimation Result on the Prosthesis walking*

Then, the proposed method is evaluated on data collected in upstair walking of the prosthesis. The estimated toe position of the prosthesis in sagittal plane and the ground truth are shown in Fig. 6(a). The average ATE of the toe in upper stair walking is 4.14 cm ± 0.85 cm (x) and 2.73 cm ± 0.71 cm (z).

In the prosthesis walking experiment, the quality of the point cloud is affected by a more drastic movement of the camera, which affects the accuracy of the prosthesis motion estimation. However, despite the slight increase in estimation errors compared to that of normal upstair walking, the system can still track the motion of the prosthesis accurately. In addition, the camera is unable to observe the stair in a few frames in prosthesis walking occasionally, by combining the inertial information, the camera position can still be estimated, which improves the robustness of the proposed method. Besides, Fig. 6(b) shows the motion of the prosthesis during walking and the constructed map after smoothing, which indicates the proposed method is still able to provide pose estimation of the prosthesis within the environment in prosthesis walking.

IV. DISCUSSION

In this study, a visual-inertial motion estimation method is proposed to determine the movement of the prosthesis within the environment. Our aim is to enhance the safety and predictability of prosthesis mobility on uneven terrains like stairs and obstacles, thereby enhancing the prosthesis's environmental adaptability. The estimation results on dataset collected during walking of different terrains and with a powered prosthesis indicate that demonstrate the method's accuracy, with an ATE of less than 5 cm in all tested terrains. Besides, the prosthetic upstair walking experiment shows that our method can be integrated with predictive control of the prosthesis and is expected to help amputees to walk on uneven terrain. More experiments on amputee walking in uneven terrain will be described in our future work.

In addition, in this study, since our main concern is the camera and prosthesis motion in the sagittal plane, the computation effort required for point cloud alignment is significantly reduced with environmental data preprocessing. According to [20], the time cost of directly performing ICP alignment for about 38k points in 3D space is about 150 ms, while the visual estimation time for each frame in this study is only 19.4 ms, which is only 12.9% of direct ICP alignment. Besides, by incorporating a SP-ESKF which is designed towards sagittal plane movements, the inertial information in the sagittal plane is fused with the visual information to obtain a more robust motion estimation of the prosthesis.

Although the proposed method successfully realizes accurate motion estimation of the prosthesis within the environment, there are still some limitations. First, only several uneven terrains in real world are considered in this study. To enhance the environmental adaptive walking in various environments, the proposed method is required to generalize to more diverse terrains in future work. Besides, if the visual estimation fails for a rather long time, the estimation relying only on the inertial information will still present a large cumulative error. Therefore, in future work, other correction methods for the inertial information, such as zero velocity update (ZUPT) [25], can be incorporated with the proposed method.

V. Conclusion

The motion of the prosthesis within the environment is important for walking in uneven terrains. In this paper, to estimate the motion of the prosthesis in sagittal plane, a visual-inertial estimation method is developed. The proposed approach utilizes the extracted feature points from the preprocessed 2D point cloud for frame-to-frame alignment. And a sagittal plane motion-oriented error-state Kalman filter is developed to fuse the inertial and visual estimation results of the camera, which is then utilized to derive the prosthesis motion. This approach is evaluated on dataset collected in walking of different terrains and in upstair walking with a powered prosthesis. The estimation result shows that the position of the prosthesis toe can be accurately estimated during walking (ATE: 4.14 cm ± 0.85 cm (x) and 2.73 cm ± 0.71 cm (z)). The proposed approach is expected to be integrated into predictive control for prosthesis on uneven terrains, thereby enhancing walking safety and environmental adaptability for amputees.